\documentclass[runningheads]{llncs}
\usepackage{graphicx}
\graphicspath{{figures/}{}}

\usepackage{amsmath,amssymb}
\usepackage[dvipsnames]{xcolor}
\usepackage{booktabs}
\usepackage{color}
\usepackage{multirow}
\usepackage{xspace}
\usepackage{array}
\newcolumntype{H}{>{\setbox0=\hbox\bgroup}c<{\egroup}@{}}

\usepackage{subcaption}
\def\model#1#2{{#1}{\scriptsize2}{#2}}

\newcommand{\st}[1]{}

\def\myfigspace{}
\def\mytabcapspace{}
\def\myfigcapspace{}
\def\mypar#1{{\smallskip \noindent\bf #1.}}

\def\fig#1{Fig.~\ref{fig:#1}}
\def\tab#1{Tab.~\ref{tab:#1}}

\makeatletter
\DeclareRobustCommand\onedot{\futurelet\@let@token\@onedot}
\def\@onedot{\ifx\@let@token.\else.\null\fi\xspace}
\def\eg{\emph{e.g}\onedot} 
\def\ie{\emph{i.e}\onedot} 
 
\def\etc{\emph{etc}\onedot} \def\vs{\emph{vs}\onedot}
\def\wrt{w.r.t\onedot} 
\def\etal{\emph{et al}\onedot}
\def \wo{\emph{w.o.}\xspace}
\makeatother

\newcommand{\omitme}[1]{}
\def\maskrcnn{{Mask R-CNN}\xspace}
\def\warp{\emph{Warp}\xspace}
\def\shift{\emph{Shift}\xspace}

\def\ftfour{Mask \model{H}{F}\xspace}

\usepackage[export]{adjustbox}
\usepackage{tabu}

\usepackage[pagebackref=false,breaklinks=true,colorlinks,bookmarks=false]{hyperref}

\begin{document}
\pagestyle{headings}
\mainmatter
\def\ECCV18SubNumber{883} 

\title{Predicting Future Instance Segmentation\\
  by Forecasting Convolutional Features}

\author{
Pauline Luc$^{1,2}$ \and
Camille Couprie$^1$ \and
Yann LeCun$^{1,3}$ \and
Jakob Verbeek$^2$
}

\institute{
Facebook AI Research\\
\and Univ. Grenoble Alpes, Inria, CNRS, Grenoble INP\thanks{Institute of Engineering Univ. Grenoble Alpes}, LJK, 38000 Grenoble, France \\
\and
New York University
}

\titlerunning{Predicting Future Instance Segmentation}

\authorrunning{Luc, Couprie, LeCun and Verbeek}

\maketitle

\begin{abstract}
Anticipating future events is an important prerequisite towards intelligent
behavior.
Video forecasting has been studied as a proxy task towards this goal.
Recent work has shown that to predict semantic segmentation of future frames, forecasting at the semantic level is more effective than forecasting RGB frames and then segmenting these.
In this paper we consider the more challenging problem of future instance
segmentation, which additionally segments out individual objects.
To deal with a varying number of output labels per image, we develop a predictive model in the space of fixed-sized convolutional features of the \maskrcnn instance segmentation model.
We apply  the ``detection head'' of \maskrcnn
on the  predicted features to produce the instance segmentation
of future frames.  Experiments show that this approach
significantly improves over  strong baselines based on optical flow and repurposed  instance segmentation architectures.
\end{abstract}

\section{Introduction}

\begin{figure}[ht]
		\def\myfig#1{\includegraphics[width = 0.435 \textwidth, valign=m]{#1}}
		{
		\begin{center}

		\newcolumntype{A}{ >{\centering\arraybackslash}m{3.76cm} }
		\newcolumntype{B}{ >{\centering\arraybackslash}m{5.36cm} }
		\begin{tabu}{B B}
			\myfig{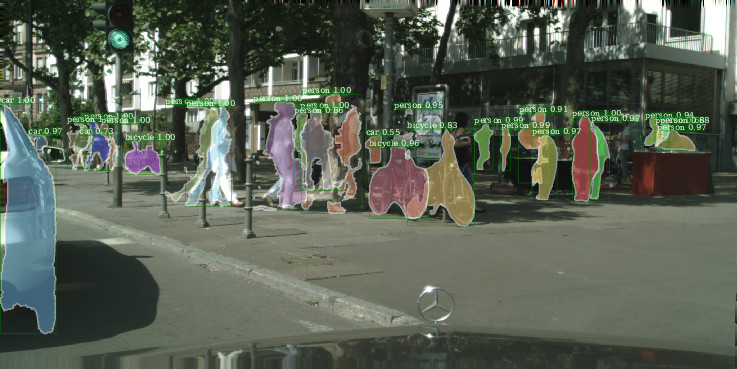} &
			\myfig{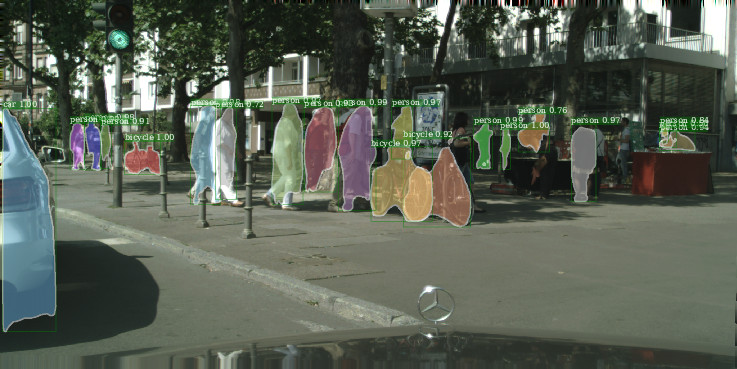}\\
			(a) optical flow baseline forecasting & {(b) our
                          instance segmentation}\\
			\myfig{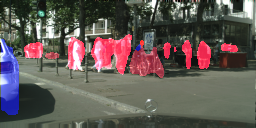} &
			\myfig{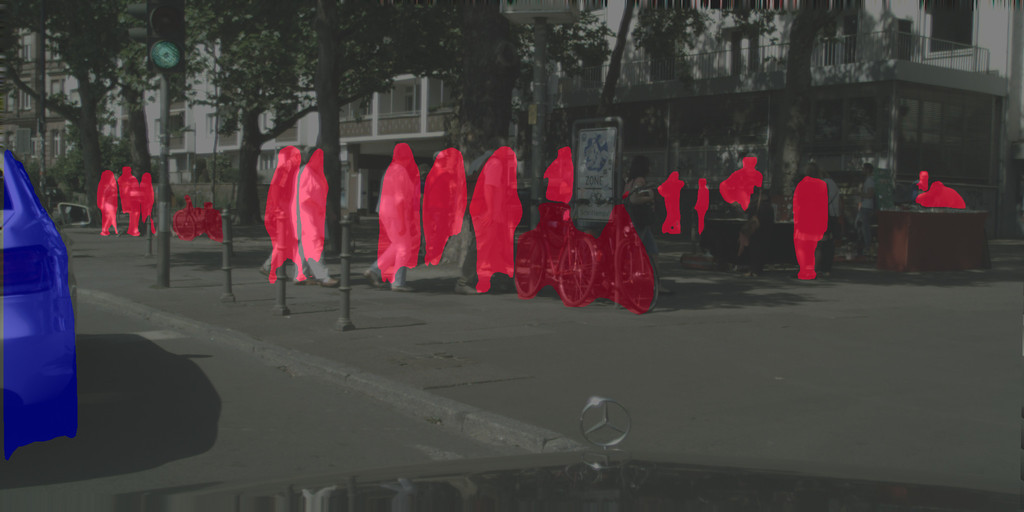} \\
(c) semantic segmentation from~\cite{luc17iccv} & {(d) our
                          semantic segmentation}
		\end{tabu}
		\end{center}
	}
    \myfigcapspace
    \caption{Predicting 0.5 sec.\ into the future. Instance modeling significantly improves the segmentation accuracy of the individual pedestrians.
	}
	\myfigspace
	\label{fig:comp_s2s_f2f}
\end{figure}

The ability to anticipate future events is a key factor towards developing intelligent behavior \cite{sutton98book}.
Video prediction  has been studied as a proxy task towards pursuing this ability, which can capitalize on the huge amount of available unlabeled video to learn visual representations that account for object  interactions and interactions between objects and the environment~\cite{mathieu16iclr}.
Most work in video prediction has focused on predicting the RGB values of future video frames~\cite{mathieu16iclr,RanzatoSzlamBruna2014,srivastava15icml,kalchbrenner17icml}.

Predictive models  have  important applications in decision-making contexts, such as autonomous driving, where rapid control decisions can be of vital importance \cite{Shalev-ShwartzB16longterm,ShalevShwartzS16sample}.
In such contexts, however, the goal is not to predict the raw RGB values of future video frames, but to make predictions about future video frames at a  semantically meaningful level, \eg in terms of presence and location of object categories in a scene.
Luc \etal \cite{luc17iccv} recently showed that for prediction of future semantic segmentation, modeling at the semantic level is much more effective than predicting raw RGB values of future frames, and then feeding these to a semantic segmentation model.

Although spatially detailed, semantic segmentation  does  not account for individual  objects, but rather lumps them together by assigning them to the same category label, \eg the pedestrians in \fig{comp_s2s_f2f}(c).
Instance segmentation overcomes this shortcoming by additionally associating
with each pixel an instance label, as show in \fig{comp_s2s_f2f}(b).
This additional level of detail is crucial for down-stream tasks that rely on instance-level trajectories, such as encountered in control for autonomous driving.
Moreover,  ignoring the notion of object instances prohibits by construction any reasoning
about object motion, deformation, \etc
Including it in the model can therefore greatly improve its predictive performance, by keeping track of individual object properties, \emph{c.f.}\ \fig{comp_s2s_f2f} (c) and (d).

Since the instance labels vary in number across frames, and do not have a consistent interpretation across videos, the approach of Luc \etal \cite{luc17iccv} does not apply to this task.
Instead, we build upon \maskrcnn~\cite{he17iccv}, a recent state-of-the-art instance segmentation model that extends an object detection system by predicting with each object bounding box a binary segmentation mask of the object.
In order to forecast the instance-level labels in a coherent manner, we predict the fixed-sized high level convolutional features used by \maskrcnn.
We obtain the future object instance segmentation by applying the \maskrcnn ``detection head'' to the predicted features.

Our approach offers several advantages:
(i) we handle cases in which the model output has a variable size, as in object detection and instance segmentation,
(ii) we do not require labeled video sequences for training, as the intermediate CNN feature maps can be computed directly from unlabeled data, and
(iii) we support models that are able to produce multiple scene interpretations, such as surface normals, object bounding boxes, and human part
labels \cite{kokkinos17cvpr}, without having to design appropriate encoders and loss functions for all these tasks to drive the future prediction.
Our contributions are the following:
\begin{itemize}
  \setlength\itemsep{0em}
\item the introduction of the new task of future instance segmentation,  which is semantically richer than previously studied anticipated recognition tasks,
\item a self-supervised approach based on predicting high dimensional CNN features of future frames,  which can support many anticipated recognition tasks,
\item experimental results that show that our feature learning approach
improves over strong baselines, relying on optical flow and  repurposed instance segmentation architectures.
\end{itemize}

\section{Related Work}

\mypar{Future video prediction}
Predictive modeling of future RGB video frames has recently been studied using a variety of techniques, including autoregressive  models \cite{kalchbrenner17icml},
adversarial training \cite{mathieu16iclr}, and recurrent networks \cite{RanzatoSzlamBruna2014,srivastava15icml,villegas17iclr}.
Villegas \etal \cite{villegas17icml} predict future human poses as a proxy to guide the prediction of future RGB video frames.
Instead of predicting  RGB values, Walker \etal \cite{walker16eccv} predict future pixel trajectories from static images.

Future prediction of more abstract representations has been considered in a variety of contexts in the past.
Lan \etal \cite{lan14eccv} predict future human actions from automatically detected atomic actions.
Kitani \etal \cite{kitani12eccv} predict future trajectories of people from semantic segmentation of an observed video frame, modeling potential destinations and transitory areas that are preferred or avoided.
Lee \etal predict future object trajectories from past object tracks and object interactions \cite{lee17cvpr}.
Dosovitskiy \& Koltun \cite{dosovitskiy17iclr} learn control models by predicting future high-level measurements in which the goal of an agent can be expressed from past video frames and measurements.

Vondrick \etal \cite{vondrick16cvpr} were the first to predict high level CNN features of future video frames to anticipate actions and object appearances in video.
Their work is similar in spirit to ours, but while they only predict image-level labels, we consider the more complex task of predicting future instance segmentation, requiring fine spatial detail.
To this end, we forecast spatially dense convolutional features, where Vondrick \etal were predicting the activations of much more compact fully connected CNN layers. Our work demonstrates the scalability of CNN feature prediction, from 4K-dimensional to 32M-dimensional features, and yields results with a surprising level of accuracy and spatial detail.

Luc \etal \cite{luc17iccv} predicted future semantic segmentation in video by taking the softmax pre-activations of past frames as input, and predicting the softmax pre-activations of future frames.
While their approach is relevant for future semantic segmentation, where the softmax pre-activations provide a natural fixed-sized representation,
it does not extend to \st{the case of} instance segmentation since the
instance-level labels vary in number between frames and are not consistent
across video sequences.
To overcome this limitation, we develop predictive models for fixed-sized convolutional features, instead of making predictions directly in the label space. Our feature-based approach has many advantages over \cite{luc17iccv}: segmenting individual instances, working at a higher resolution and providing a framework that generalizes to other dense prediction tasks.
In a direction orthogonal to our work, Jin \etal \cite{jin17nips}
jointly predict semantic segmentation and optical flow of future frames, leveraging the complementarity between the two tasks.

\mypar{Instance segmentation approaches}
Our approach can be used in conjunction with any deep network to perform instance segmentation.
A variety of approaches for instance segmentation has been explored in the
past, including iterative object segmentation using recurrent networks
\cite{romera16eccv}, watershed transformation \cite{bai17cvpr}, and object
proposals \cite{pinheiro16eccv}.  In our work we build upon \maskrcnn~\cite{he17iccv}, which recently established a new state-of-the-art for
instance segmentation.
This method extends the Faster R-CNN object detector~\cite{ren15nips} by adding a network branch to
predict segmentation masks and extracting features for prediction in a way that
allows precise alignment of the masks when they are stitched together to form
the final output.

\section{Predicting Features for Future Instance Segmentation}

In this section we briefly review the \maskrcnn instance segmentation framework, and then present how we can use it for anticipated recognition by predicting internal CNN features of future frames.

\subsection{Instance Segmentation with \maskrcnn}
\label{sec:maskrcnn}

The \maskrcnn model~\cite{he17iccv} consists of three main stages.  First, a
convolutional neural network (CNN) ``backbone'' architecture is used to extract
high level feature maps.  Second, a region proposal network (RPN) takes these
features to produce regions of interest (ROIs), in the form of coordinates of
bounding boxes susceptible of containing instances.  The bounding box proposals
are used as input to a \emph{RoIAlign} layer, which interpolates the high level
features in each bounding box to extract a fixed-sized representation for each
box. Third, the features of each RoI are
input to the detection branches, which produce refined bounding box coordinates,
a class prediction, and a fixed-sized binary mask for the predicted class.
Finally, the mask is interpolated back to full image resolution within the
predicted bounding box and reported as an instance segmentation for the
predicted class. We refer to the combination of the second and third stages as
the ``detection head''.

\begin{figure}[t]
\begin{center}
\includegraphics[width=1\linewidth]{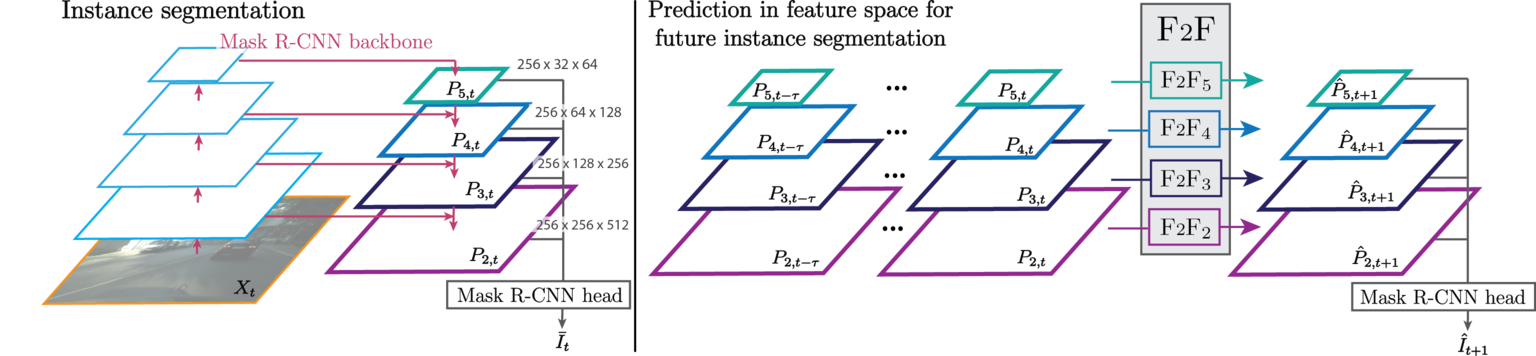}
\end{center}
\myfigcapspace
   \caption{Left: Features in the FPN backbone are obtained by upsampling
     features in the top-down path, and combining them with features from the
     bottom-up path at the same resolution.
     Right: For future      instance segmentation, we extract FPN features from frames $t-\tau$ to $t$,
     and predict the FPN features for frame $t+1$.  We learn separate
     feature-to-feature prediction models for each FPN level: \model{F}{F}$_l$
     denotes the model for level $l$.
      }
   \label{fig:approach_f2f}
   \myfigcapspace
\end{figure}

He \etal \cite{he17iccv} use a Feature Pyramid Network (FPN) \cite{lin17cvpr} as backbone architecture, which extracts a set of features at several spatial resolutions from an input image. The feature pyramid is then used in the instance segmentation pipeline to detect objects at multiple scales, by running the detection head on each level of the pyramid.
Following \cite{lin17cvpr}, we denote the feature pyramid levels extracted from an RGB image $X$ by P$_2$ through P$_5$, which are of decreasing resolution $(H/2^{l} \times W/2^{l})$ for P$_l$, where $H$ and $W$ are respectively the height and width of $X$.
The features in P$_l$ are computed in a top-down stream by up-sampling those in P$_{l+1}$ and adding the result of a 1$\times$1 convolution of features in a layer with matching resolution in a bottom-up ResNet stream.
We refer the reader to the left panel of \fig{approach_f2f} for a schematic illustration, and to \cite{he17iccv,lin17cvpr} for more details.

\subsection{Forecasting Convolutional Features}
\label{sec:future}

Given a video sequence, our goal is to predict instance-level object segmentations for one or more future frames, \ie for frames where we cannot access the RGB pixel values.
Similar to previous work that predicts future RGB frames  \cite{mathieu16iclr,RanzatoSzlamBruna2014,srivastava15icml,kalchbrenner17icml} and future semantic segmentations \cite{luc17iccv}, we are interested in models where the input and output of the predictive model live in the same space, so that the model can be applied recursively to produce predictions for more than one frame ahead.
The instance segmentations themselves, however, do not provide a suitable representation for prediction,  since  the instance-level labels vary in number between frames, and are not consistent across video sequences.
To overcome this issue, we instead resort to predicting the highest level features in the \maskrcnn architecture that are of fixed size.
In particular, using the FPN backbone in \maskrcnn, we want to learn a model that
given the  feature pyramids extracted from  frames $X_{t-\tau}$ to $X_t$,
predicts the feature pyramid for the unobserved RGB frame $X_{t+1}$.

\mypar{Architecture}
The features at the different FPN levels are trained to be input to a shared detection head, and are thus of similar nature.
However, since the resolution changes across levels, the spatio-temporal dynamics are distinct from one level to another.
Therefore, we propose a multi-scale approach, employing a separate network to
predict the features at each level, of which we demonstrate the benefits in
  Section \ref{sec:setup}.
The per-level networks are trained and function completely independently from each other. This allows us to parallelize the training across multiple GPUs. 
Alternative architectures in which prediction across different resolutions is tied are interesting, but beyond the scope of this paper.
For each level, we concatenate the features of the input sequence along the feature dimension.
We refer to the ``feature to feature'' predictive model for level $l$ as \model{F}{F}$_l$.
The overall architecture is summarized in the right panel of  \fig{approach_f2f}.

Each of the \model{F}{F}$_l$ networks is implemented by a resolution-preserving CNN.
Each network is itself multi-scale as in \cite{mathieu16iclr,luc17iccv},
to efficiently enlarge the field of view while preserving high-resolution
details.
More precisely, for a given level $l$, \model{F}{F}$_l$ consists of $s_l$ subnetworks
\model{F}{F}$_l^s$, where $s \in \{1,...,s_l \}$.
The network \model{F}{F}$_l^{s_l}$ first processes the input downsampled by a factor of $2^{s_l -1}$.
Its output is up-sampled by a factor of $2$,
and concatenated to the input downsampled by a factor of  $2^{s_l -2}$.
This concatenation constitutes the input of \model{F}{F}$_l^{s_l-1}$ which predicts a refinement of the initial coarse prediction.
The same procedure is repeated until the final scale subnetwork \model{F}{F}$_l^1$.
The design of subnetworks \model{F}{F}$_l^s$ is inspired by \cite{luc17iccv}, leveraging dilated convolutions to further enlarge the
field of view.
Our architecture differs  in the number of feature maps per layer, the convolution kernel sizes and dilation parameters, to make it more suited for the larger input dimension. We detail these design choices in the supplementary material.

\mypar{Training}
We first train the \model{F}{F}$_5$ model to predict the coarsest features P$_5$, precomputed offline. Since the features of the different FPN levels are fed to the same recognition head network, the next levels are similar to the P$_5$ features. Hence, we initialize the weights of \model{F}{F}$_4$, \model{F}{F}$_3$, and \model{F}{F}$_2$ with the ones learned by \model{F}{F}$_5$,
before fine-tuning them. For this, we compute features on the fly, due to memory constraints. Each of the \model{F}{F}$_l$ networks is trained using an $\ell_2$ loss.

For multiple time step prediction, we can fine-tune each subnetwork
\model{F}{F}$_l$ autoregressively using backpropagation through time, similar to
\cite{luc17iccv} to take into account error accumulation over time. In this case, given a single sequence of input feature
maps, we train with a separate $\ell_2$ loss on each predicted future frame. In
our experiments, all models are trained in this autoregressive manner, unless
specified otherwise.

\section{Experimental Evaluation}

In this section we first present our experimental setup and baseline models, and
then proceed with quantitative and qualitative results, that demonstrate the strengths of our \model{F}{F} approach.

\subsection{Experimental Setup: Dataset and Evaluation Metrics}
\label{sec:setup}

\mypar{Dataset}
In our experiments, we use the Cityscapes dataset \cite{cordts16cvpr} which contains 2,975 train, 500 validation and 1,525 test video sequences of 1.8 second each, recorded from a car driving in urban environments.
Each sequence consists of 30 frames of resolution 1024$\times$2048. Ground truth semantic and instance segmentation annotations are available for the 20-th frame of each sequence.

We employ a \maskrcnn model pre-trained on the MS-COCO dataset \cite{lin14eccv} and fine-tune it in an end-to-end fashion on the Cityscapes dataset, using a ResNet-50-FPN backbone.
The coarsest FPN level P5 has resolution 32$\times$64, and the finest level P2 has resolution 256$\times$512. 

Following \cite{luc17iccv}, we temporally subsample the videos by a factor three, 
and take four frames as input.
That is, the input sequence consists of feature pyramids for frames $\{X_{t-9},
X_{t-6}, X_{t-3}, X_{t}\}$.
We denote by \emph{short-term} and \emph{mid-term} prediction respectively predicting $X_{t+3}$ only (0.17 sec.) and through $X_{t+9}$ (0.5 sec.). We additionally evaluate \emph{long-term} predictions, corresponding to $X_{t+27}$ and 1.6 sec.\ ahead on the two long Frankfurt  sequences of the Cityscapes validation set.

\mypar{Conversion to semantic segmentation}
For direct comparison to previous work, we also convert our instance segmentation predictions to  semantic segmentation.
To this end, we first assign to all pixels the \emph{background} label.
Then, we iterate over the detected object instances in order of ascending confidence score.
For each instance, consisting of a confidence score $c$, a class $k$, and a binary mask $m$, we either reject it if it is lower than a threshold $\theta$ and accept it otherwise, where in our experiments we set $\theta=0.5$.
For accepted instances, we update the spatial positions corresponding to mask $m$ with label $k$.
This step potentially replaces labels set by instances with lower confidence,
and resolves competing class predictions.

\mypar{Evaluation  metrics}
To measure the instance segmentation performance, we use the standard Cityscapes metrics.
The average precision metric AP50 counts an instance as correct if it has at least 50\% of intersection-over-union (IoU) with the ground truth instance it has been matched with.
The summary AP metric is given by average AP obtained with ten equally spaced IoU thresholds from 50\% to 95\%.
Performance is measured across the eight classes with  available instance-level ground truth:
\emph{person, rider, car, truck, bus, train, motorcycle,} and \emph{bicycle}.

We measure semantic segmentation performance across the same eight classes. In addition to the IoU metric, computed \wrt the ground
truth segmentation of the 20-th frame in each sequence, we also quantify the
segmentation accuracy using three standard segmentation measures used in
\cite{yang08cviu}, namely the Probabilistic Rand Index (RI)
\cite{pantofaru05tr}, Global Consistency Error (GCE) \cite{martin01iccv}, and
Variation of Information (VoI) \cite{meila05icml}. Good segmentation results are
associated with high RI, low GCE and low VoI.

\mypar{Implementation details and ablation study} We cross-validate the number of scales, the
optimization algorithm and hyperparameters per level of the pyramid. 
For each level of the pyramid a single scale network was selected, except for
\model{F}{F}$_2$, where we employ 3 scales. The \model{F}{F}$_5$ network is
trained for 60K iterations of SGD with Nesterov Momentum of $0.9$,
learning rate $0.01$, and batch size of 4 images.
It is used to initialize the other networks, which are trained for 80K
iterations of SGD with Nesterov Momentum of $0.9$, batch size of 1 image and learning rates of $5\times 10^{-3}$ for \model{F}{F}$_4$ and $0.01$ for \model{F}{F}$_3$. For \model{F}{F}$_2$, which is much deeper, we used Adam with learning rate $5\times 10^{-5}$ and default parameters.
Table~\ref{tab:ablation_study} shows the positive
  impact of using each additional feature level, denoted by P$_i$--P$_5$ for
  $i=2,3,4$. We also report performance when using all features levels,
  predicted by a model trained on the coarsest P$_5$ features, shared across
  levels, denoted by P$_5$ //. The drop in performance \wrt the column
  P$_2$-P$_5$ underlines the importance of training specific networks for each
  feature level.

\begin{table}[t]
\begin{center}
{\small
\setlength{\tabcolsep}{6pt}
\begin{tabular}{lccccc}
\toprule
Levels &  P$_5$ &  P$_4$--P$_5$ &  P$_3$--P$_5$ &  P$_2$--P$_5$ & P$_5$ // \\
\midrule
IoU   & 15.5  & 38.5 & 54.7 & \bf{60.7} & 38.7\\ 
AP50 & 2.2 & 10.2 & 24.8 & \bf{40.2}  & 16.7\\
\bottomrule
\end{tabular}}
\end{center}
\caption{Ablation study: short-term prediction on the Cityscapes val.\ set.}
\label{tab:ablation_study}
\myfigcapspace
\end{table}

\subsection{Baseline Models}

As a performance upper bound, we report the accuracy of a \maskrcnn oracle that has access to the future RGB frame.
As a lower bound, we also use a trivial copy baseline that returns the segmentation of the last input RGB frame. Besides the following baselines, we also experiment with two weaker baselines, based on nearest neighbor search and on predicting the future RGB frames, and then segmenting them. We detail both baselines in the supplementary material.

\mypar{Optical flow baselines}
We designed two baselines using the optical flow field computed from the last input RGB frame to the second last, as well as the instance segmentation
predicted at the last input frame.
The \warp approach consists in warping each instance mask independently using the flow field inside this mask.
We initialize a separate flow field for each instance, equal to the flow field inside the instance mask and zero elsewhere.
For a given instance, the corresponding flow field is used to project the values of the instance mask in the opposite direction of the flow vectors, yielding a new binary mask. To this predicted mask, we associate the class and confidence score of the input instance it was obtained from.
To predict more than one time-step ahead, we also update the instance's flow field in the same fashion, to take into account the previously predicted displacement of physical points composing the instance. The predicted mask and flow field are used to make the next prediction, and so on.
Maintaining separate flow fields allows competing flow values to coexist for the same spatial position, when they belong to different instances whose predicted trajectories lead them to overlap.
To smoothen the results of this baseline, we perform post-processing operations at each time step, which significantly improve the results and which we detail in the supplementary material.

Warping the flow field when predicting multiple steps ahead suffers from error accumulation.
 To avoid this,  we test another baseline, \shift,  which shifts each mask with the average
flow vector computed across the mask.
To predict $T$ time steps ahead, we simply shift the instance $T$ times.
This approach, however, is unable to scale the objects, and is therefore unsuitable for long-term prediction when objects significantly change in  scale as their distance to the camera changes.

\mypar{Future semantic segmentation using discrete label maps}
For comparison with the future semantic segmentation approach of \cite{luc17iccv},
which ignores instance-level labels, we train their \model{S}{S} model on the label maps produced by \maskrcnn.
Following their approach, we down-sample the \maskrcnn label maps to $128 \times 256$.
Unlike the soft label maps from the Dilated-10 network \cite{yu16iclr} used in \cite{luc17iccv}, our converted \maskrcnn label maps are discrete.
For autoregressive prediction, we discretize the output by replacing the softmax
network output with a one-hot encoding of the most likely class at each position.
For autoregressive fine-tuning, we use a softmax activation with a low
temperature parameter at the output of the \model{S}{S} model, to produce
near-one-hot probability maps in a differentiable way, enabling backpropagation through time.

\mypar{Future segmentation using the Mask R-CNN architecture} As another baseline, we fine-tune
  Mask R-CNN to predict mid-term future segmentation given the last 4 observed
  frames, denoted as the \ftfour baseline. As initialization, we replicate the weights of the first layer learned on the COCO dataset across the 4 frames, and divide
  them by 4 to keep the features at the same scale.

\begin{table*}[tb]
\mytabcapspace

\begin{center}
{\small
\setlength{\tabcolsep}{6pt}
\begin{tabular}{lcccc}
\toprule
& \multicolumn{2}{c}{Short-term} & \multicolumn{2}{c}{Mid-term} \\
& AP50 & AP & AP50 & AP \\
\midrule
\maskrcnn oracle & 65.8 & 37.3 & 65.8 & 37.3 \\
\midrule
Copy last segmentation  &  24.1   &  10.1   &  6.6  &  1.8 \\
Optical flow -- \shift   &  37.0   &  16.0   &  9.7  &  2.9 \\
Optical flow -- \warp &  36.8   &   16.5  &  11.1  & 4.1 \\
\ftfour* &  25.5 & 11.8 & 14.2 & 5.1 \\
\midrule
\model{F}{F} w/o ar.\ fine tuning & \bf{40.2} & 19.0 & 17.5 & 6.2\\
\model{F}{F}  & 39.9 & \bf{19.4} & \bf{19.4} & \bf{7.7}\\
\bottomrule
\end{tabular}
}
\end{center}
\caption{Instance segmentation accuracy on the Cityscapes validation set.
\mbox{* Separate} models were trained for short-term and mid-term predictions.}
\label{tab:main_res_inst}
\end{table*}

\subsection{Quantitative Results}
\label{sec:results}

\mypar{Future instance segmentation} In \tab{main_res_inst} we present instance
segmentation results of our future feature prediction approach (\model{F}{F})
and compare it to the performance of the oracle, copy, optical flow and
  \ftfour baselines. The copy baseline performs
very poorly (24.1\% in terms of AP50 \vs 65.8\% for the oracle), which
underlines the difficulty of the task.  
The two optical flow baselines perform comparably for short-term prediction, and are both much 
better than the copy baseline.
For mid-term prediction, the \warp approach outperforms \shift.
The \ftfour
  baseline performs poorly for short-term prediction, but its results degrade
  slower with the number of time steps predicted, and it outperforms the \warp
  baseline for mid-term prediction. As \ftfour outputs a single
  time step prediction, either for short or mid-term predictions, it is not
  subject to accumulation of errors, but each prediction setting requires training a specific model. Our \model{F}{F} approach gives the best results overall, reaching
more than 37\% of relative improvement over our best mid-term baseline.
While our \model{F}{F} autoregressive fine-tuning makes little difference in
case of short-term prediction (40.2\% \vs 39.9\% AP50 respectively), it gives a
significant improvement for mid-term prediction (17.5\% \vs 19.4\% AP50
respectively).

\begin{figure*}[htb]
  \myfigspace
\def\myfig#1{\includegraphics[width = 0.24 \textwidth,
    valign=m]{#1}}
\def\myfigl#1{\includegraphics[width = 0.22 \textwidth,
    valign=m]{#1}}
\def\myfigu#1{\includegraphics[width = 0.88 \textwidth,
    valign=m]{#1}}
\begin{tabular}{cc}
\begin{minipage}{0.8\textwidth}
\begin{subfigure}[b]{1\textwidth}
       {\tabulinesep=0.1mm
		\newcolumntype{A}{ >{\centering
		}b{23mm} }
		\begin{tabu}{AAAA}
		\myfig{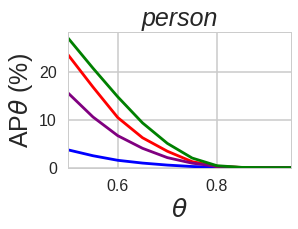}&
		\myfig{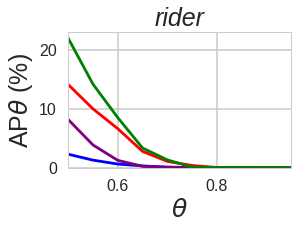}&
		\myfig{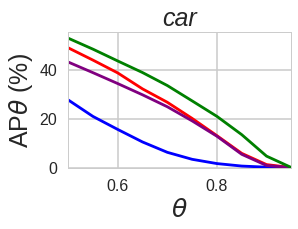} &
		\myfig{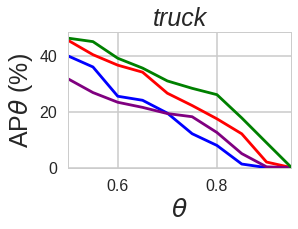}\\
		\myfig{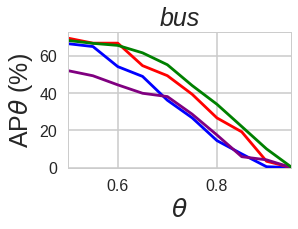} &
		\myfig{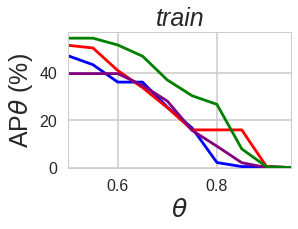} &
		\myfig{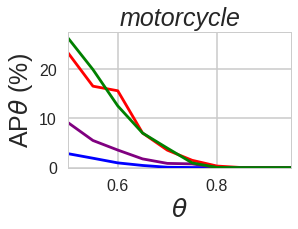}&
		\myfig{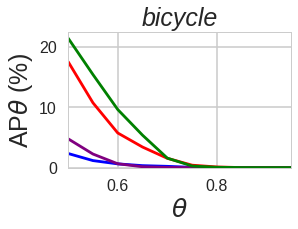} \\
		\myfigl{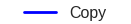} &
		\myfigl{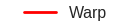} &
		\myfigl{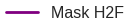}&
		\myfigl{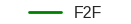} \\
		\end{tabu}
	}
        \caption{Short-term - Individual classes scores}
        \label{fig:quant_results_f2f_pc_pt_short:a}
\end{subfigure}
\end{minipage}&
\begin{minipage}{0.22\textwidth}
\begin{subfigure}[b]{1\textwidth}
       {\tabulinesep=0.1mm
		\newcolumntype{A}{ >{\centering
		}b{23mm} }
		\begin{tabu}{A}
		\myfigu{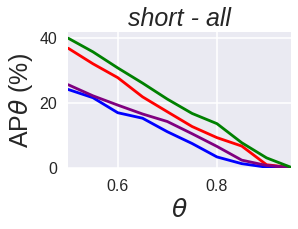} \\
		\myfigu{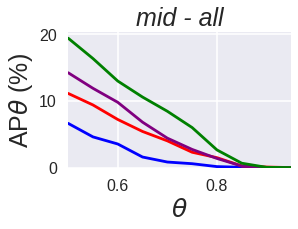} \\
		\myfigl{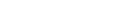}
		\end{tabu}
	}
        \caption{Overall}
        \label{fig:quant_results_f2f_pc_pt_short:b}
\end{subfigure}
\end{minipage}
\end{tabular}
	\caption{Instance segmentation AP$\theta$ across different IoU thresholds $\theta$.
	(a) Short-term prediction per class; (b) Average across all classes for
          short-term (top) and mid-term prediction (bottom).}
		\label{fig:quant_results_f2f_pc_pt_short}
		\myfigspace
\end{figure*}

In \fig{quant_results_f2f_pc_pt_short}(a), we show how the AP metric varies with the IoU threshold, for short-term prediction across the different classes and for each method.
For individual classes, \model{F}{F} gives the best results across thresholds, except for very few exceptions.
In \fig{quant_results_f2f_pc_pt_short}(b), we show average results over all classes for short-term and mid-term prediction. We see that \model{F}{F} consistently improves over the baselines across all thresholds, particularly for mid-term prediction.

\begin{table}[t]
\begin{center}
{\small
\setlength{\tabcolsep}{6pt}
\begin{tabular}{lcccccccccc}
\toprule
 && \multicolumn{4}{c}{Short-term}  && \multicolumn{4}{c}{Mid-term} \\
 && IoU  &  RI & VoI & GCE && IoU &  RI & VoI & GCE\\
\midrule
Oracle  \cite{luc17iccv} && 64.7 & --- & --- & --- && 64.7 & --- & --- & --- \\
\model{S}{S} \cite{luc17iccv} && 55.3  & --- & --- & --- && 40.8 & --- & --- & ---\\
\midrule
Oracle && 73.3 & 94.0 & 20.8 & 2.3 && 73.3 & 94.0 & 20.8 & 2.3\\
\midrule
Copy && 45.7 & 92.2 & 29.0 & 3.5 && 29.1 & 90.6 & 33.8 & 4.2 \\
\shift && 56.7 & 92.9 & 25.5 & 2.9 && 36.7 & 91.1 & 30.5 & 3.3 \\
\warp && 58.8 & {\bf 93.1} & 25.2 & 3.0&& 41.4 & 91.5 & 31.0 &  3.8 \\
\ftfour* && 46.2  & 92.5 & 27.3 & 3.2 &   & 30.5 & 91.2 & 31.9 & 3.7 \\
\model{S}{S}  && 55.4 & 92.8 & 25.8 & 2.9 && {\bf 42.4} & 91.8 & 29.7 & 3.4\\
\model{F}{F}  && {\bf   61.2}& {\bf 93.1}& {\bf 24.8} & {\bf 2.8} && 41.2 & {\bf 91.9} & {\bf 28.8} & \bf 3.1 \\
\bottomrule
\end{tabular}}
\end{center}
\caption{Short and mid-term semantic segmentation of moving objects (8 classes) performance on the Cityscapes validation  set.
\mbox{* Separate} models were trained for short-term and mid-term predictions.} 
\label{tab:sem_segm_res}
\end{table}

\mypar{Future semantic segmentation} We additionally provide a comparative
evaluation on semantic segmentation in \tab{sem_segm_res}.  First, we observe
that our discrete implementation of the \model{S}{S} model performs slightly
better than the best results obtained by \cite{luc17iccv}, thanks to our better
underlying segmentation model (\maskrcnn \vs the Dilation-10 model
\cite{yu16iclr}). Second, we see that the \ftfour baseline performs weakly
  in terms of semantic segmentation metrics for both short and mid-term
  prediction, especially in terms of IoU. This may be due to frequently
  duplicated predictions for a given instance, see
  Section \ref{sec:qual}. Third, the advantage of \warp over \shift appears
clearly again, with a 5\% boost in mid-term IoU.  Finally, we find that
\model{F}{F} obtains clear improvements in IoU over all methods for short-term
segmentation, ranking first with an IoU of 61.2\%. Our \model{F}{F} mid-term IoU is comparable to those of
the \model{S}{S} and \warp baseline, while being much more accurate in depicting
contours of the objects as shown by consistently better RI, VoI and GCE
segmentation scores.

\begin{figure*}[htb]
    \def\myfig#1{\includegraphics[width = 0.305 \textwidth, valign=m]{#1}}
    {\tabulinesep=0.4mm
    \begin{center}

    \newcolumntype{A}{ >{\centering\arraybackslash}m{3.76cm} }
    \newcolumntype{B}{ >{\centering\arraybackslash}m{3.7cm} }
    \begin{tabu}{c B B B}
& {\centering \warp} &{\centering \ftfour} & {\centering \model{F}{F}}\\
	\rotatebox[origin=c]{90}{(1)} &
      \myfig{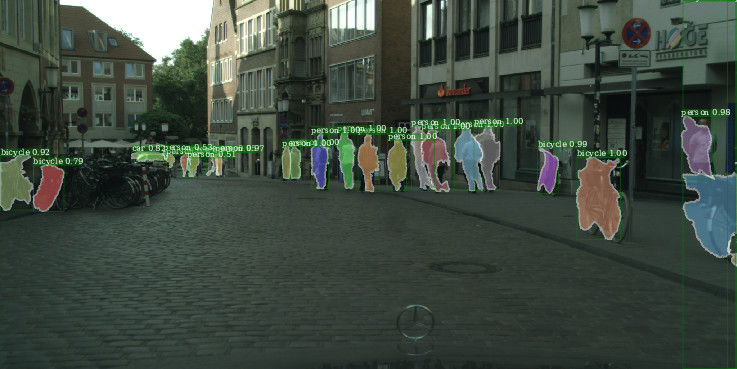} &
      \myfig{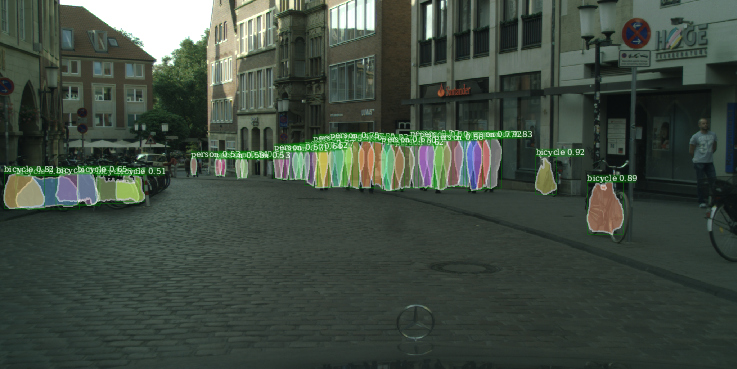} & 
      \myfig{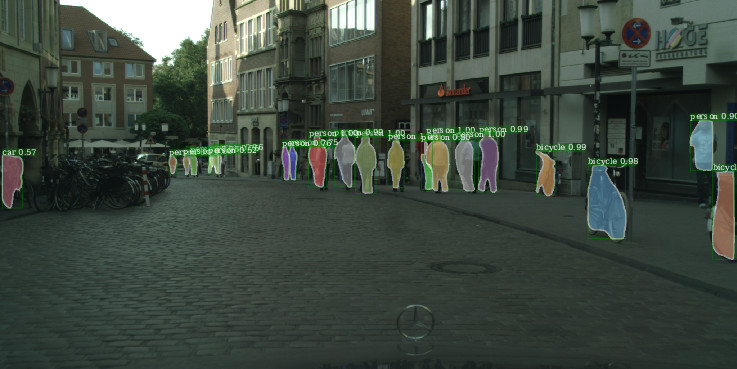} \\
	\rotatebox[origin=c]{90}{(2)} &
      \myfig{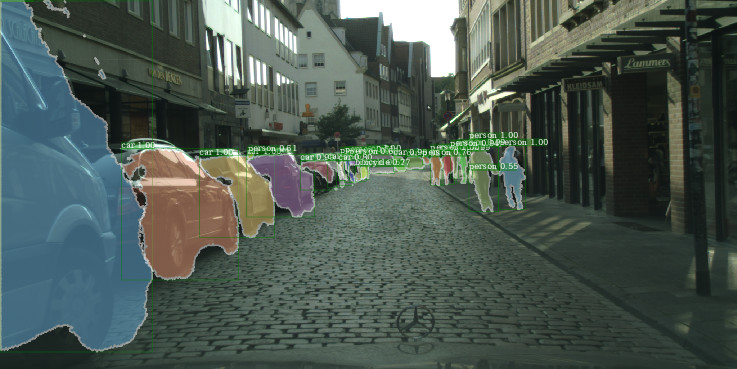}&
      \myfig{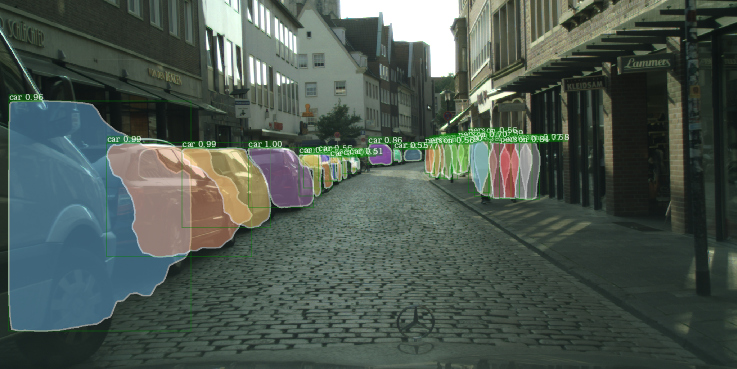}& 
      \myfig{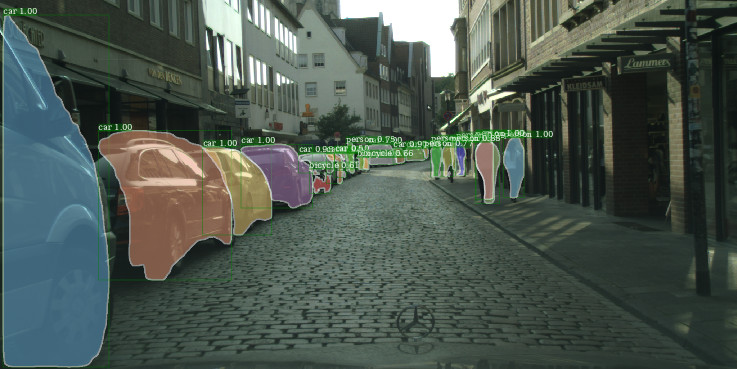} \\
	\rotatebox[origin=c]{90}{(3)} &
      \myfig{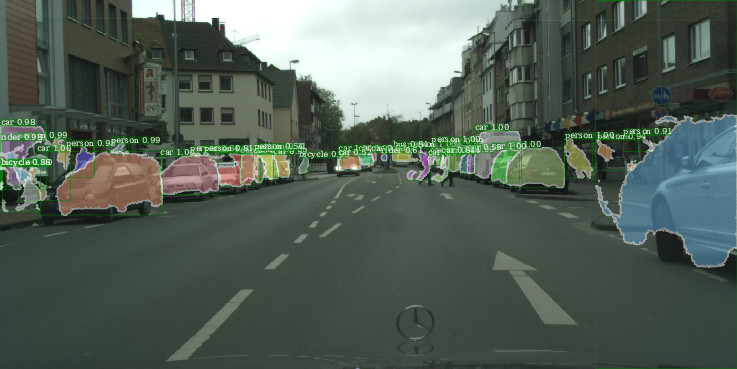} &
      \myfig{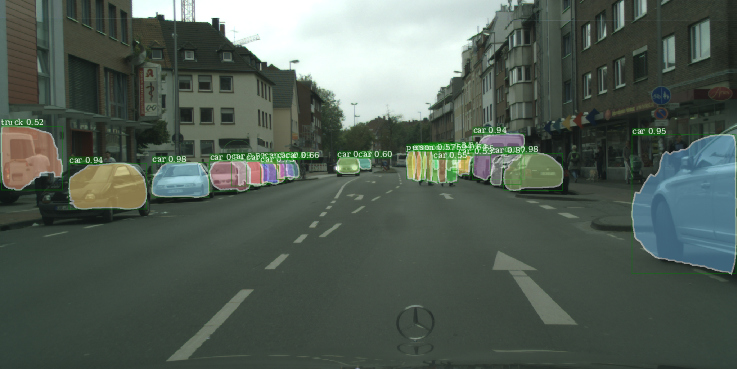} & 
      \myfig{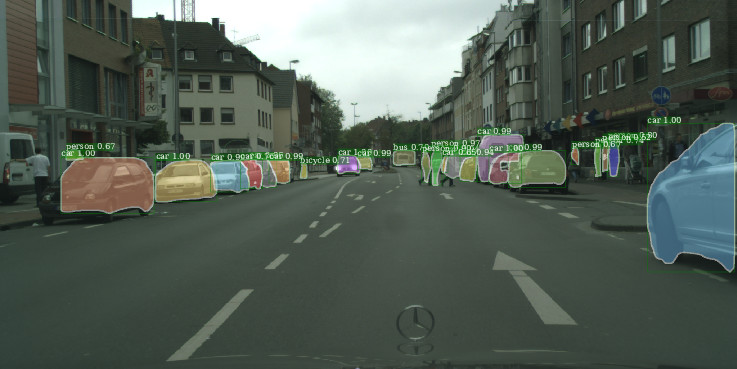} \\
    \end{tabu}
    \end{center}
  }
  \caption{Mid-term instance segmentation predictions (0.5 sec.\ future)
      for 3 sequences, from left to right: \warp baseline,
      \ftfour baseline and \model{F}{F}.}
    \label{fig:qual_results_f2f_inst}
\end{figure*}

\subsection{Qualitative Results}
\label{sec:qual}
\begin{figure*}[htb]
    \def\myfig#1{\includegraphics[width = 0.305 \textwidth, valign=m]{#1}}
    {\tabulinesep=0.4mm
    \begin{center}

    \newcolumntype{A}{ >{\centering\arraybackslash}m{3.76cm} }
    \newcolumntype{B}{ >{\centering\arraybackslash}m{3.7cm} }
    \begin{tabu}{c B B B}
    & {\centering (1)} &{\centering (2)} & {\centering (3)}\\
    \rotatebox[origin=c]{90}{\model{S}{S}} &
      \myfig{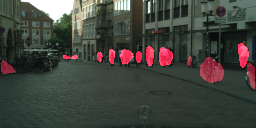} &
      \myfig{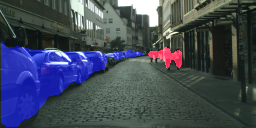} &
      \myfig{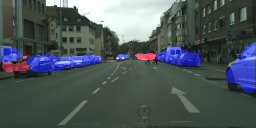} \\
     \rotatebox[origin=c]{90}{\model{F}{F}} &
      \myfig{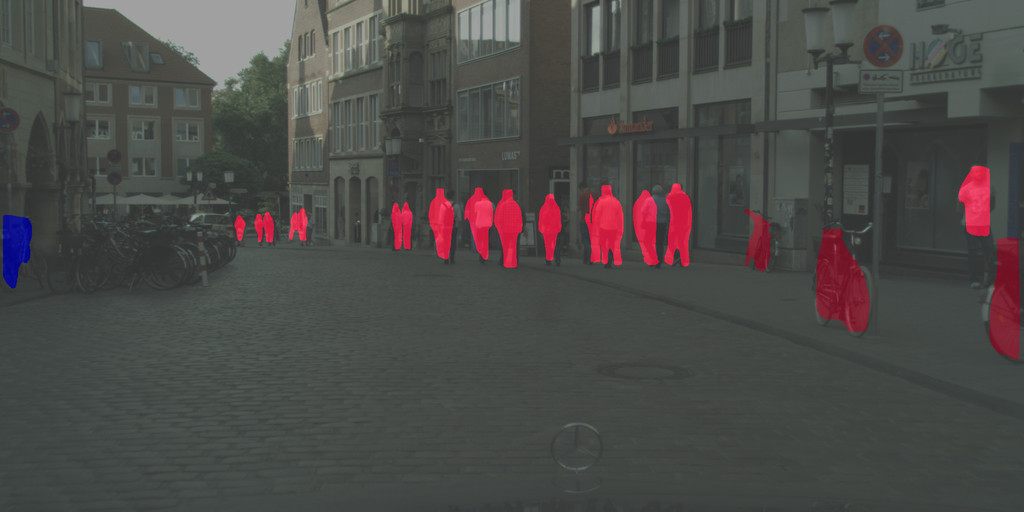} &
      \myfig{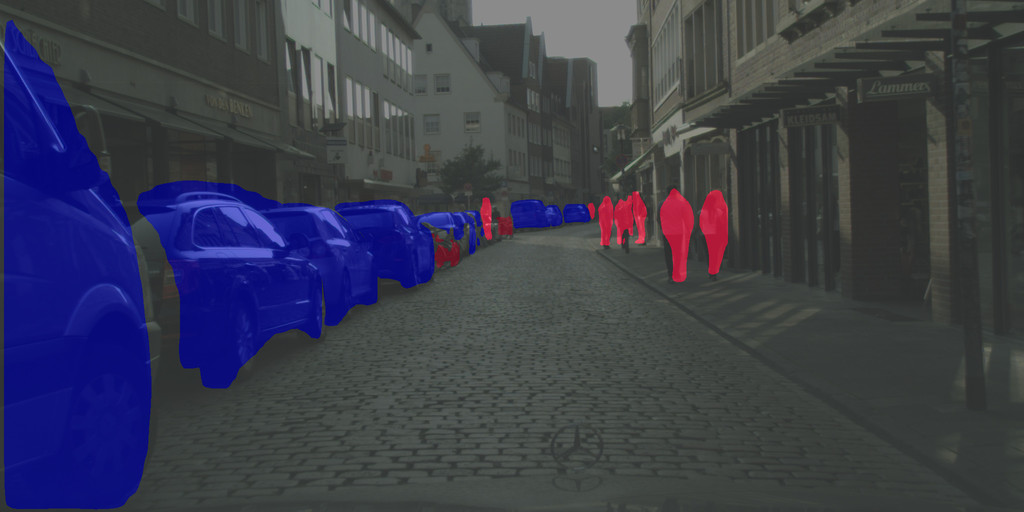}&
      \myfig{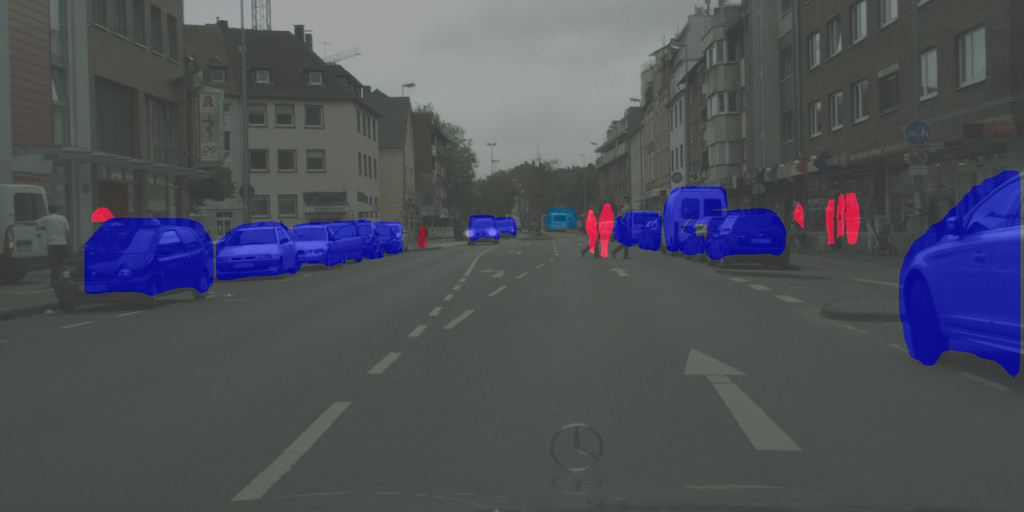} \\
    \end{tabu}
    \end{center}
  }
    \myfigcapspace
  \caption{Mid-term semantic segmentation predictions (0.5 sec.) for 3
      sequences. For each case we show from top to bottom: \model{S}{S} model
      and \model{F}{F} model.}
    \label{fig:qual_results_f2f_sem}
\end{figure*}

Figures \ref{fig:qual_results_f2f_inst} and \ref{fig:qual_results_f2f_sem} show representative results of our approach, both in
terms of instance and semantic segmentation prediction, as well as results from
the \emph{Warp} and \ftfour baselines for instance segmentation and \model{S}{S} for semantic segmentation. We visualize predictions with a threshold of 0.5 on the confidence of masks.
The \ftfour baseline frequently predicts several masks around objects, especially for objects with ambiguous trajectories, like pedestrians, and less so for more predictable categories like cars. We speculate that this is due to the loss that the network is optimizing, which does not discourage this behavior, and due to which the network is learning to predict several plausible future positions, as long as they overlap sufficiently with the ground-truth position. This does not occur with the other methods, which are either optimizing a per-pixel loss or are not learned at all.
\model{F}{F} results are often better aligned with the actual layouts of the
objects than the \emph{Warp} baseline,
showing that our approach has learned to model dynamics of the scene and objects more accurately than the baseline.
As expected, the predicted masks are also much more precise than those of the
\model{S}{S} model, which is not instance-aware.

\begin{figure*}[h!]
	\def\myfig#1{\includegraphics[width = 0.305 \textwidth]{#1}}
	\begin{center}
			\begin{tabular}{cc}
			\warp & \model{F}{F}\\
				\myfig{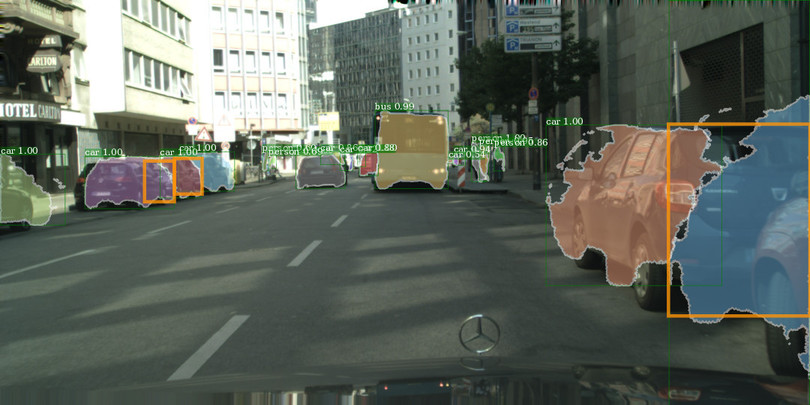} &
				\myfig{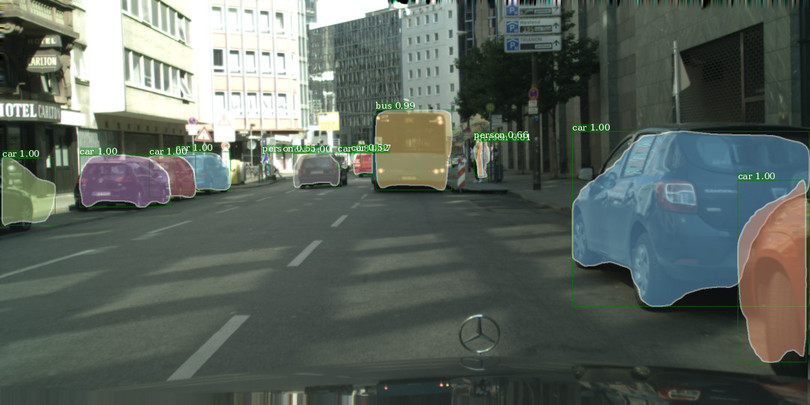}
			\end{tabular} \\
			\begin{tabular}{cc}
				\myfig{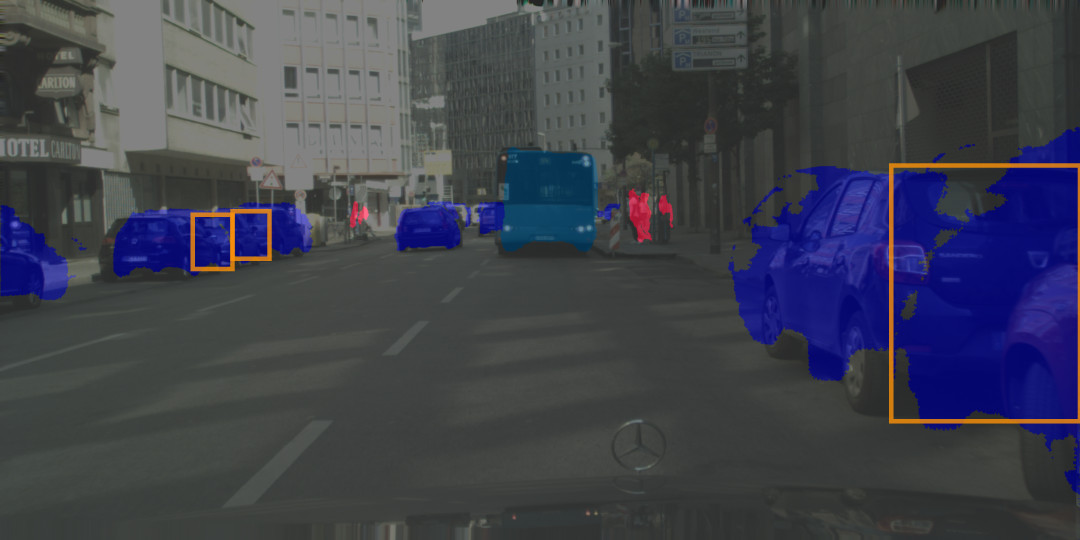} &
				\myfig{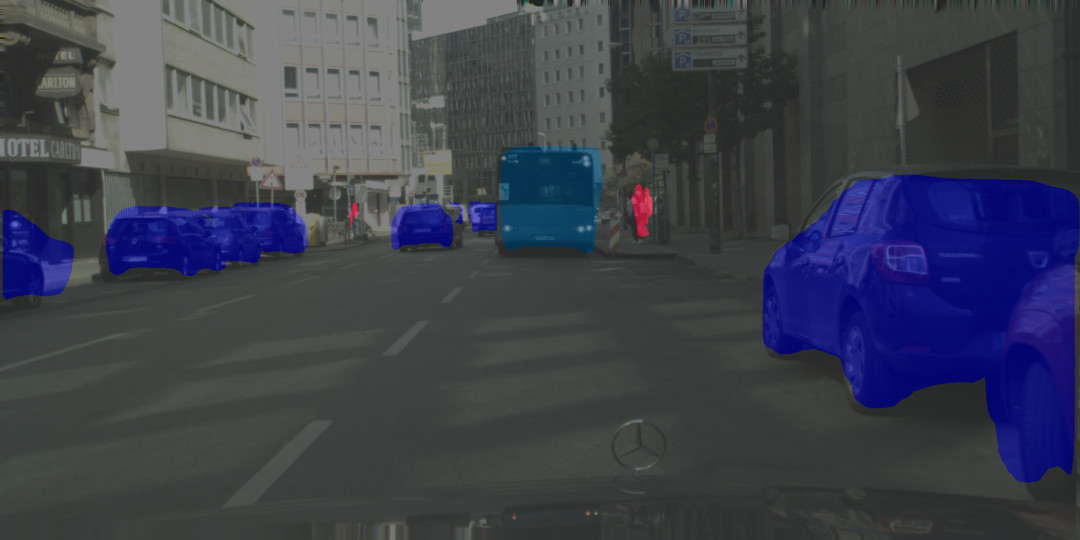}
			\end{tabular} \\
	\end{center}
\caption{Mid-term predictions of instance and semantic segmentation with the \emph{Warp} baseline and our \model{F}{F} model.
Inaccurate instance segmentations can result in accurate semantic segmentation areas; see orange rectangle highlights.
		}
		\label{fig:comp_warp_f2f}

\end{figure*}

In \fig{comp_warp_f2f} we provide additional examples to better understand why
the difference between \model{F}{F} and the \warp baseline is smaller for
semantic segmentation metrics than for instance segmentation metrics.  When
several instances of the same class are close together, inaccurate estimation of
the instance masks may still give acceptable semantic segmentation.  This
typically happens for groups of pedestrians and rows of parked cars. If an
instance mask is split across multiple objects, this will further affect the AP
measure than the IoU metric.  The same example also illustrates common artifacts
of the \warp baseline that are due to error accumulation in the propagation of
the flow field.

\begin{figure*}[htb]
		\def\myfig#1{\includegraphics[width = 0.305\textwidth]{#1}}
		\begin{center}
			\begin{tabular}{cc}
				\myfig{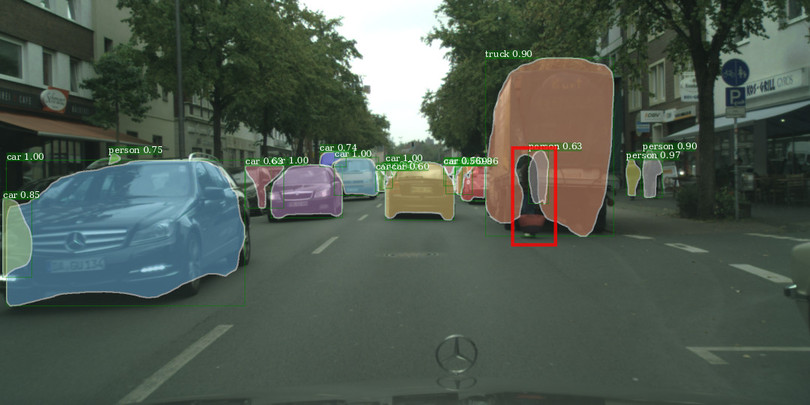} &
				\myfig{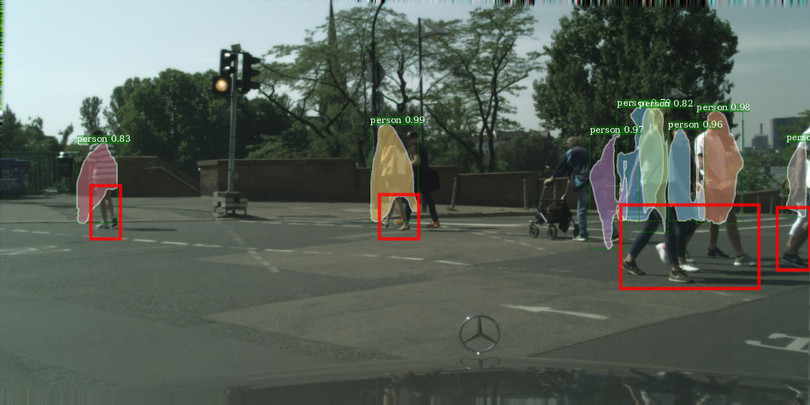} \\
				(a) & (b) \\
			\end{tabular}
		\end{center}
    \myfigcapspace
	\caption{Failure modes of mid-term prediction with the \model{F}{F} model, highlighted with the red boxes:
	incoherent masks (a), lack of detail in highly deformable object regions, such as legs of pedestrians (b).
	}
	\myfigcapspace
		\label{fig:qual-failure-cases}
\end{figure*}

\subsection{Discussion}

\mypar{Failure cases} To illustrate some of the remaining challenges in
predicting future instance segmentation we present several failure cases of our
\model{F}{F} model in \fig{qual-failure-cases}. 
In \fig{qual-failure-cases}(a), the masks predicted for the truck and the person
are incoherent, both in shape and location. More consistent predictions might be
obtained with a mechanism for explicitly modeling occlusions.
Certain motions and shape transformations are hard to predict
accurately due to the inherent ambiguity in the problem.  This is, \eg, the case
for the legs of pedestrians in \fig{qual-failure-cases}(b), for which there is a
high degree of uncertainty on the exact pose.  Since the model is deterministic,
it predicts a rough mask due to averaging over several possibilities.  This may
be addressed by modeling the intrinsic variability using GANs, VAEs, or
autoregressive models~\cite{kalchbrenner17icml,goodfellow14nips,kingma14iclr}.

\begin{figure*}[htb]
\begin{center}
	\includegraphics[width=0.92\textwidth]{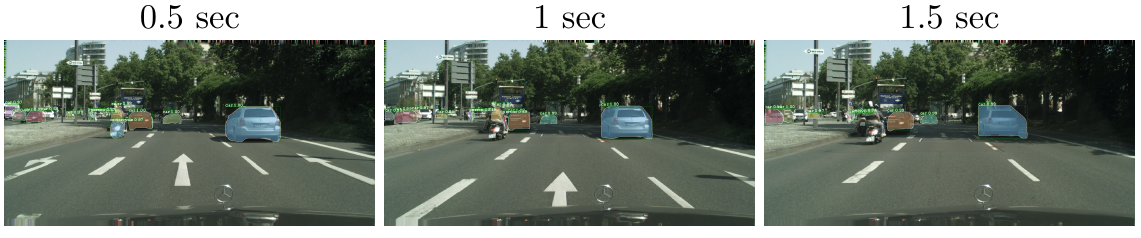}
\end{center}
	\caption{Long-term predictions (1.5 seconds) from our \model{F}{F}
          model.}
        \label{fig:long}
\end{figure*}

\mypar{Long term prediction} In \fig{long} we show a prediction of \model{F}{F} 
up to 1.5 sec. in the future in a sequence of the long Frankfurt video of the
Cityscapes validation set, where frames were extracted with an interval of 3 as
before.  To allow more temporal consistency between predicted objects, we apply
an adapted version of the method of Gkioxari \etal \cite{gkioxari15cvpr} as a
post-processing step. We define the linking score as the sum of confidence
scores of subsequent instances and of their IoU. We then greedily compute the paths between instances which maximize these scores using the Viterbi algorithm.  We thereby
obtain object tracks along the (unseen) future video frames.  Some object
trajectories are forecasted reasonably well up to a second, such as the rider, while others are lost by that time such as the motorbike.  We also compute the AP with the ground truth of the long Frankfurt video. For each method, we give the best result of 
    either predicting 9 frames with a frame interval of 3, or the opposite. For \ftfour, only the latter is possible, as there are no such long sequences available for training.
 	We obtain an AP of 0.5 for the flow and copy baseline, 0.7 for \model{F}{F} and 1.5 for \ftfour. 
    All methods lead to very low scores, highlighting the severe
    challenges posed by this problem.

\section{Conclusion}

We introduced a new anticipated recognition task: predicting instance segmentation of future video frames.
This task is defined at a semantically meaningful level rather the level of raw RGB values, and
adds instance-level information as compared to predicting future semantic segmentation.
We proposed a generic and self-supervised approach for anticipated recognition based on predicting the convolutional features of future video frames.
In our experiments we apply this approach in combination with the \maskrcnn
instance segmentation model.  We predict the internal ``backbone'' features
which are of fixed dimension, and apply the ``detection head'' on these
features to produce a variable number of predictions.  Our results
show that future instance segmentation can be predicted much better than
naively copying the segmentations from the last observed frame, and that our
future feature prediction approach significantly outperforms two strong baselines, the first one relying on optical-flow-based warping and the second on repurposing and fine-tuning the \maskrcnn architecture for the task.  When evaluated on the more
basic task of semantic segmentation without instance-level detail, our approach
yields performance quantitatively comparable to earlier approaches, while having
qualitative advantages.

Our work shows that with a  feed-forward network we are able to obtain surprisingly accurate results. More sophisticated architectures have the potential to further improve performance. Predictions may be also improved by explicitly modeling the temporal consistency of instance segmentation, and predicting multiple possible futures rather than a single one.

We invite the reader to watch videos of our predictions at \url{http://thoth.inrialpes.fr/people/pluc/instpred2018}.

\medskip
{{\bf Acknowledgment.}
This work has been partially supported by the grant ANR-16-CE23-0006 ``Deep in
France'' and LabEx PERSYVAL-Lab (ANR-11-LABX-0025-01). We thank Matthijs Douze, Xavier Martin, Ilija Radosavovic and Thomas Lucas for their precious comments.
}

\bibliographystyle{splncs}
\bibliography{luc2018instpred}

\appendix

\section{Future instance segmentation results on the test set}

We provide instance segmentation results on the Cityscapes test set in Table~\ref{tab:test_inst} for mid-term prediction, as obtained from the online evaluation server.
For reference, we also computed the \maskrcnn oracle results (prediction using the future RGB 
frame), and the results of baselines \warp and \ftfour. 
The results are comparable to those on the validation set, and we again observe that the results of  our \model{F}{F}  model are  far more accurate than those of the baselines.

\begin{table*}[h]
\mytabcapspace

\begin{center}
{\small
\setlength{\tabcolsep}{6pt}
\begin{tabular}{lcc}
\toprule
& \multicolumn{2}{c}{Mid-term}\\
&  AP50 & AP \\
\midrule
\maskrcnn oracle & 58.1  & 31.9  \\
\midrule
Optical flow -- \warp &  11.8  &  4.3   \\ 
\ftfour &  12.2  &  4.6   \\ 
\model{F}{F}  & \bf{17.5} & \bf{6.7} \\ 
\bottomrule
\end{tabular}
}
\end{center}
\caption{Mid-term instance segmentation results on the Cityscapes test set}
\myfigspace
\label{tab:test_inst}
\end{table*}

\section{Details on optical flow baselines}

To obtain the optical flow estimates, we employed the Full Flow method \cite{chen16cvpr} using the default parameters given by the authors on the MPI Sintel Flow Dataset.

\subsection{Ablation study for the post-processing on \warp}

\begin{figure*}[h!]
	\centering
	\includegraphics[width=0.9\textwidth]{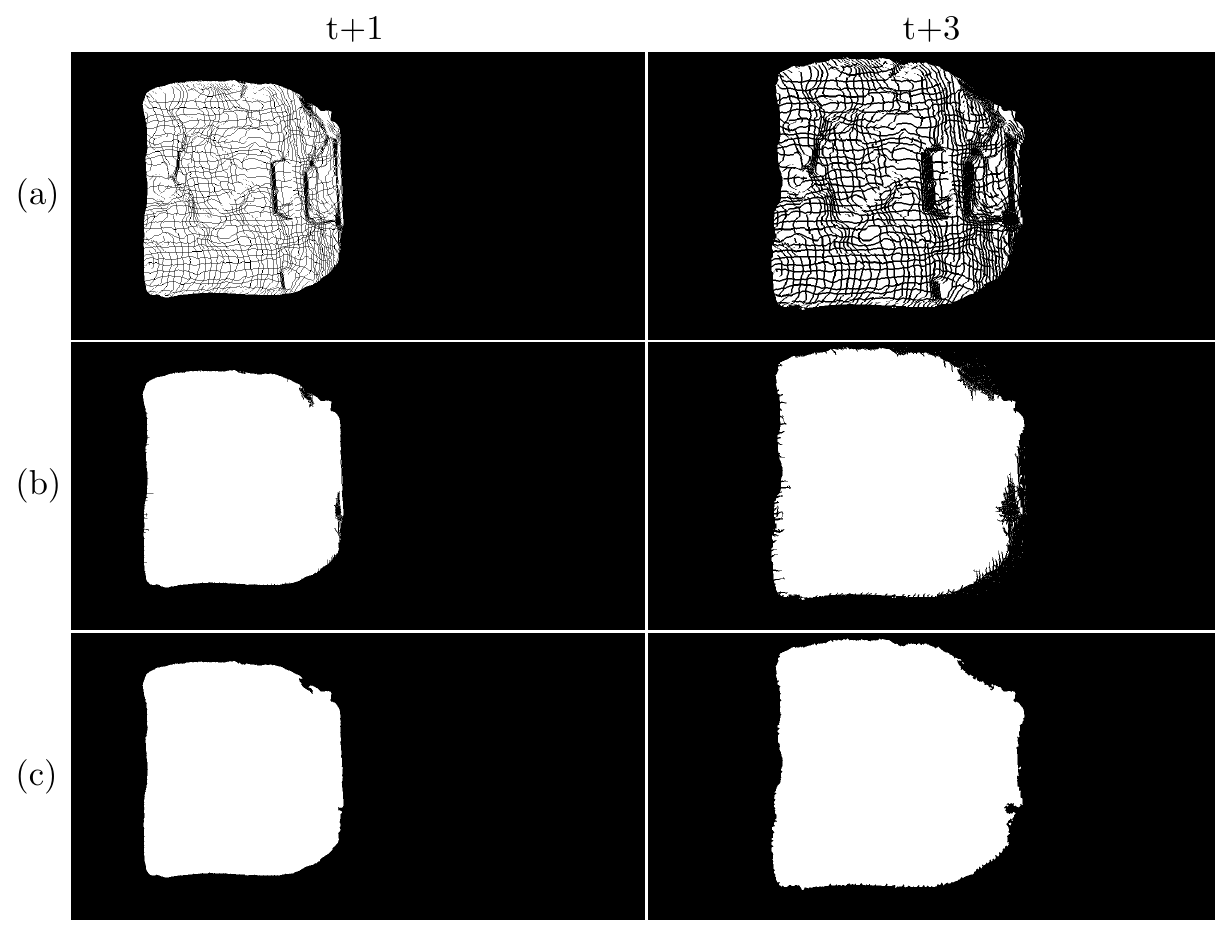}
	\caption{Qualitative comparison of masks obtained using the \warp approach: (a) \wo post-processing, (b) with closing operations, and (c) with full post-processing. }
	\label{fig:qual_ablation}
	\myfigspace
\end{figure*}

Prior to any post-processing, the \warp baseline predictions present some artifacts, as shown in \fig{qual_ablation}(a), in particular when objects are moving fast towards the camera. 
In this case, the optical flow should lead the predicted mask to become larger.
But by construction, the number of pixels composing the masks can only stay equal or decrease in
the warping process.
Masks are therefore broken in parts corresponding to uniform areas of the flow field,
and this phenomenon worsens with the number of steps.

In order to remove these artifacts, we employ mathematical morphology operators to post-process the predictions. 
First we employ a morphological closing, followed by a closing of holes on the
masks.
This addresses the problem in an effective manner, as shown in \fig{qual_ablation}(b).

For mid-term predictions, we perform these operations on the output before it is used as input, at each time step.
We use bilinear interpolation to estimate flow values at the added positions of the binary
mask. This post-processing of the flow adds small spurious artifacts at the border of the masks, visible in particular in  \fig{qual_ablation}(b), right. These are easily removed using morphological openings, see \fig{qual_ablation}(c).

In \tab{ablation_flow} we report the performance of the \warp baseline corresponding to the illustrations in \fig{qual_ablation}: (a) 
 before any post-processing is applied (\warp \wo post-processing), (b) with closing operations only (\warp  \wo opening), and (c) with full post-processing (\warp). These results show that the post processing operations we employ significantly improve performance.
 
\begin{table*}[h!]
\mytabcapspace
\begin{center}
{\small
\setlength{\tabcolsep}{6pt}
\begin{tabular}{lcccH}
\toprule
& \multicolumn{4}{c}{Mid-term} \\
& AP50 & AP & IoU  & mIOU SEGM\\
\midrule
\warp \wo post-processing &  5.7 & 1.6 & 32.2 & 33.8   \\
\warp  \wo opening &  10.9 & 4.0 & 40.6 & 42.3   \\
\warp  & 11.1 & 4.1 & 41.4 & 42.9   \\
\bottomrule
\end{tabular}
}
\end{center}
\caption{Ablation study on the Cityscapes validation set for the post-processing operations employed by the
  \warp optical flow baseline. 
  }
  \myfigspace
\label{tab:ablation_flow}
\end{table*}

\subsection{Qualitative comparison with \shift}

\begin{figure*}[h!]
	\centering
	\includegraphics[width=\textwidth]{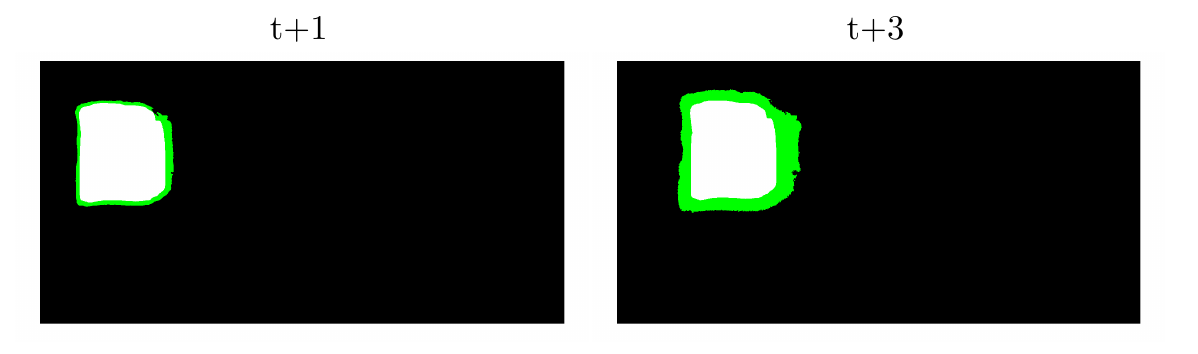}
	\caption{Comparison between the masks predicted by \shift, in white, and \warp, the union of the white and green zones. Predictions are shown for short and mid-term.}
	\label{fig:comp-shift-warp}
	\myfigcapspace
\end{figure*}

The \shift optical flow baseline leads to qualitatively better masks in cases where the optical flow field is not accurate enough. This approach, however, is unable to scale the objects. We illustrate this in \fig{comp-shift-warp}, in an example where a train is approaching the camera. At the first prediction, the mask predicted by \shift has nicer contours than that of \warp. However, one can already see that the \warp mask is a bit larger. By the third prediction, we see that this has become much more accentuated. As a consequence, \shift does not reach the performance of the \warp approach, as reported in the main paper.

Disentangling the camera motion from that of the instances and incorporating additional geometric priors to additionally scale masks might improve the results of the \shift approach, but is outside the scope of this work.

\section{Additional baselines}

In this section we present two additional baselines that were suggested by anonymous reviewers, 
based on RGB frame prediction and nearest neighbor search.

\subsection{Prediction in RGB space followed by segmentation}

Predictive modeling has applications in decision-making contexts; in such scenarios however, predictions are required to be semantically meaningful. To show that prediction in the feature space is more effective than in the RGB space, we use \maskrcnn to segment future RGB frames predicted using the \model{X}{X} model of \cite{luc17iccv}. The resulting AP50 on the validation set of the Cityscapes dataset for short-term prediction is 6.9\%, while the AP is 3.6\%. This is much weaker than even the copy baseline, which reaches 24.1\% AP50 and 10.1\% AP in the same setting. When fine-tuning \maskrcnn on the predicted (blurry) RGB frames of the training set, rather than the normal RGB frames, and keeping the same optimization hyperparameters used for fine-tuning on the original Cityscapes dataset, we obtain 19.2\% AP50 and 8.6\%, closer to but still below the copy baseline, and far from our \model{F}{F} results 40.2\% AP50 and 19.0 AP\%.

\begin{figure*}[h!]
	\centering
	\includegraphics[width=\textwidth]{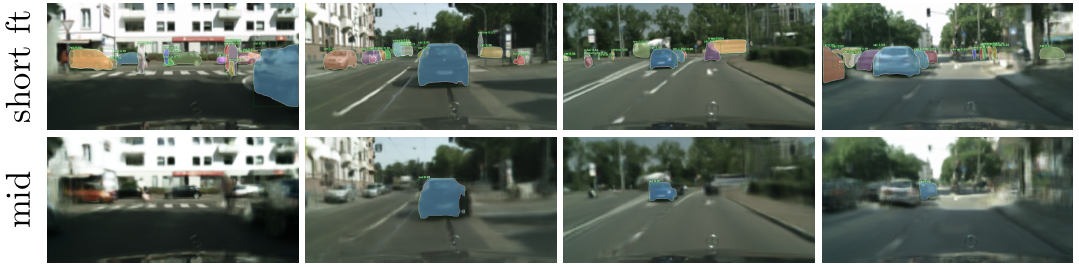}
	\caption{Predictions in RGB space, followed by instance segmentation prediction using a \maskrcnn model. Top: short-term predictions, using a \maskrcnn model fine-tuned to this setting; bottom: mid-term predictions, using the original \maskrcnn model.}
	\label{fig:x2x}
	\myfigspace
\end{figure*}

\subsection{Nearest neighbor baseline}

The nearest neighbor baseline takes the P$_5$ features of the last observed frame, finds the nearest training frame in $\ell_2$ distance on the features, and outputs the future segmentation of the matched frame. This segmentation corresponds to the ground-truth annotation if it is available, otherwise it is produced by the \maskrcnn oracle. The searching set comprises the input frames used to train \model{S}{S} and \model{F}{F}, \ie frames 2, 5, 8, 11 of each training sequence.

\begin{figure*}[h!]
	\centering
	\includegraphics[width=\textwidth]{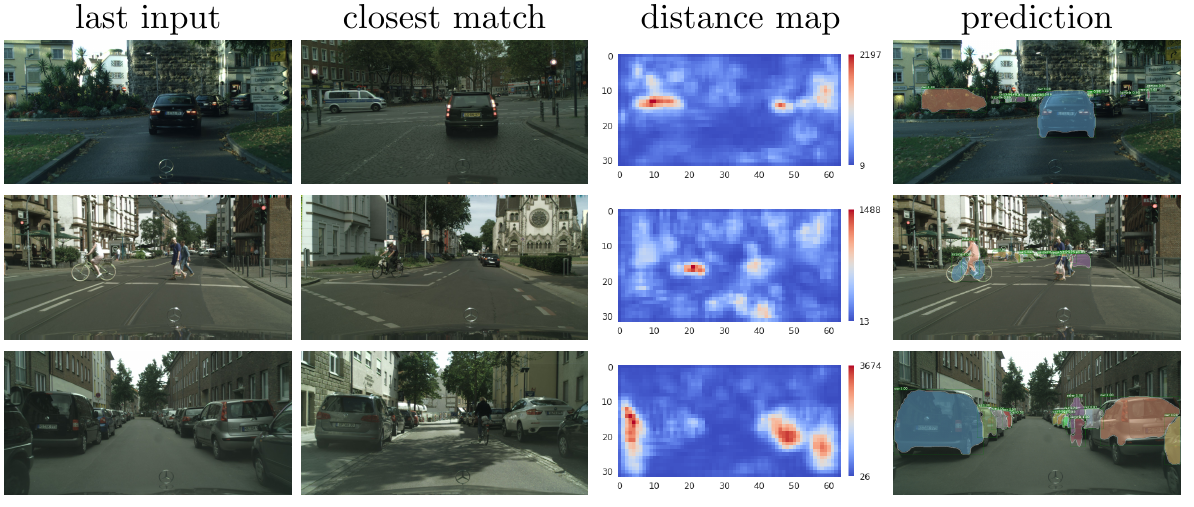}
	\caption{Nearest neighbor baseline. For each example, we show the last input frame, its closest match, the corresponding squared pixelwise $\ell_2$ distance heat map and the predicted instance segmentations, visualized over the actual future frame.}
	\label{fig:nearest_neighbours}
		\myfigcapspace
\end{figure*}

This baseline obtains very poor results, with an AP50 of 0.3\% and IoU of 7.9\%. This is due to the limited size of the dataset, and the large number of instances present in each frame: each image contains on average 7 humans and 12 vehicles~\cite{cordts16cvpr}. Although the nearest neighbor baseline sometimes accurately matches large instances, the other objects lead to a great number of false positives and false negatives, severely degrading the performance. 
We show examples where this occurs in \fig{nearest_neighbours}.

\section{\model{F}{F} architecture design}

\begin{figure}[th!]
\centering
	\begin{tabular}{c|c}
		\hspace*{0.2cm} \includegraphics[width=0.37\textwidth]{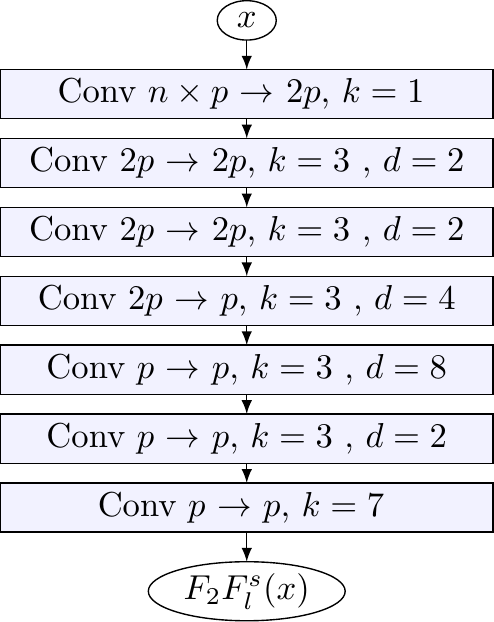}		 \hspace*{0.5cm} &
		\hspace*{0.5cm} \includegraphics[width=0.37\textwidth]{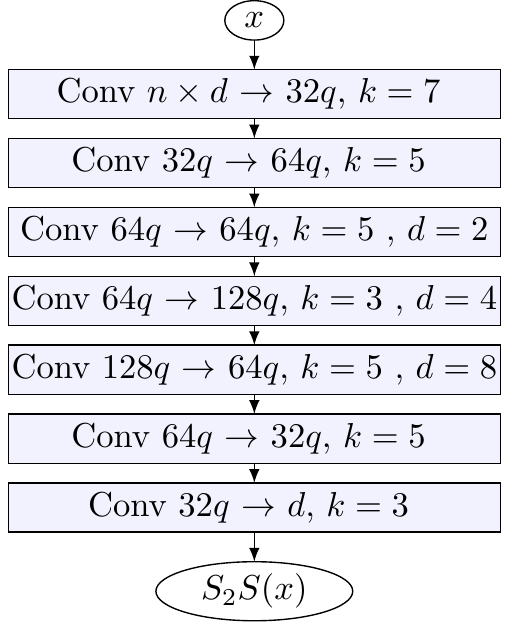}	 \hspace*{0.2cm} \\
Field of view: 43 & 		 Field of view: 65   \\
		(a) & (b) 
	\end{tabular}
	\caption{Architecture design of (a) \model{F}{F}$_{l}^s$ and (b) \model{S}{S} from \cite{luc17iccv}.
          Each convolutional layer except the final one is followed by a ReLU.
          Stride is always one, padding is chosen so as to maintain the size of
          the input. The parameter $q$ of \model{S}{S} was set to 1.5 as in \cite{luc17iccv}.
        }
	\label{fig:f2f_ls_architecture}
\end{figure}

We recall that our \model{F}{F} model is composed of four networks: \model{F}{F}$_{l}$,
where $l\in\{2, 3, 4, 5\}$, to forecast features at varying scales. Each network
may be itself multiscale and is composed in this case of $s_l$ subnetworks
\model{F}{F}$_{l}^s$, where $s \in \{1,...,s_l \}$.
Each subnetwork is fed with an input having a channel dimension $n \times p$, where $n$ is the
number of input frames, including the coarse prediction output by the previous
subnetwork, and $p$ is the channel dimension of the input and target feature
space. In our experiments we have $n=4$ (or $n=5$ including the previous coarse prediction), and $p=256$.
To ease comparison, our architecture closely follows that of \cite{luc17iccv}, modifying the number of layers and dilation parameters to scale the architecture to the high dimension of our input and target feature space.
We summarize both architectures in \fig{f2f_ls_architecture}.

\end{document}